# Physics Guided Conditional Diffusion Framework for Generative Inverse Design of Manufacturable Metasurface based Absorbers

Vineetha Joy[1*], Jamshed Palai[1], Satwik Sahu[2], Anshuman Kumar[3], Amit Sethi[3], Hema Singh[1]
[1]Centre for Electromagnetics, CSIR-National Aerospace Laboratories, Kodihalli, Bangalore 560017
[2]Birla Institute of Technology and Science, Pilani, Rajasthan 333031
[3]Indian Institute of Technology, Bombay, Maharashtra 400076
vineethajoy@nal.res.in, ompalai11@gmail.com; stwk17@gmail.com; anshuman.kumar@iitb.ac.in, asethi@iitb.ac.in, hemasingh@nal.res.in

***Abstract*— Inverse design of metasurfaces under continuous electromagnetic (EM) constraints requires generation of geometries that simultaneously satisfy stringent spectral specifications and remain manufacturable. Conventional approaches based on iterative full-wave simulations are computationally prohibitive for large design spaces, while existing generative models often suffer from poor conditional controllability and limited fabrication awareness. In this regard, we propose a physics guided condition-quality enhanced diffusion framework for the inverse design of metasurface based absorbers. Fabrication-aware constraints involving commercially available materials and feasible thicknesses are incorporated to ensure practical realizability of the generated designs. The framework introduces a conditioning mechanism for continuous spectral specifications, wherein feature-wise linear modulation (FiLM) propagates the condition across the denoising hierarchy, enabling stable and accurate generation with improved spectral controllability. Further, to embed EM consistency directly into the generative learning process, a pre-trained surrogate EM simulator is integrated within the diffusion training pipeline. The proposed framework generated physically realizable metasurface designs for diverse reflection characteristics in the frequency range of 2–18 GHz, achieving a very low average spectral mean squared error of 0.0006 and a high band alignment accuracy of 0.958. The framework also addresses the fundamentally non-unique nature of inverse EM design by enabling structured multimodal generation of geometrically distinct yet spectrally consistent metasurface designs for the same target response. The proposed model produces the suitable design in approximately 30 seconds, whereas the conventional approach can take several months under comparable computational resources. The efficiency of the model is also established via experimental measurements.**

**Keywords—*Metasurfaces, Absorbers, Diffusion models, feature-wise linear modulation* (*FiLM*)*, Conditioning Quality.***

## I. Introduction

Metasurfaces are two-dimensional, artificially engineered structures composed of subwavelength scatterers called meta-atoms [1]. By tailoring the geometry, constituent materials, and spatial arrangement of these meta-atoms, metasurfaces can precisely manipulate the amplitude, polarization, and phase of incident electromagnetic (EM) waves. If a metasurface with sub-wavelength resonators is placed at the boundary between two homogeneous media, the reflection and transmission characteristics will alter as per generalized Snell's laws of refraction and reflection [2], [3]. This happens because the metasurface induces a resonant response that modifies the phase of the wave. Hence, the direction and characteristics of the incident wavefront can be modified by varying the geometry and spatial arrangement of meta-atoms on a metasurface. Further, from the perspective of practical realization, metasurfaces can be easily fabricated and are more space efficient than their three-dimensional counterparts, i.e., metamaterials. These attractive attributes have made metasurfaces highly promising for applications such as stealth technology [4]- [6], invisibility cloaking [7], EM shielding [8], holographic imaging [9], [10], wireless communication [11], [12], lensing [13], [14], optical multiplexing [15] etc.

However, the inverse design of metasurfaces (Fig. 1) where specific EM responses must be mapped to physical structures is a very challenging task. This is due to the high dimensionality, nonlinearity, and non-uniqueness inherent in the design space. Conventional design approaches typically rely on the intuition of the designer, rigorous parametric sweeps, and iterative full-wave simulations to arrive upon a suitable design [16]-[19]. While effective for simple configurations, these methods become excessively time-consuming and computationally intensive for metasurfaces with even moderate design complexity. This process often requires hundreds or thousands of EM simulations each of which can take minutes to hours. Various optimization-based methods [20] such as topology optimization [21], [22], genetic algorithms [23]-[26], ant-colony optimization [27]-[29], particle swarm optimization (PSO) [30], [31], etc. have been developed to improve the efficiency of the above-mentioned traditional design scheme. These algorithms rely on iteratively evaluating the performance of candidate structures until a specific criterion is met. Although these approaches outperform manual trial-and-error strategies, they are still constrained by slow convergence requiring large number of iterations, high computational cost due to repeated full-wave simulations, susceptibility to local optima, and limited generalizability. Moreover, they often require complete re-optimization for each new performance specification, making them impractical for multi-objective design tasks.

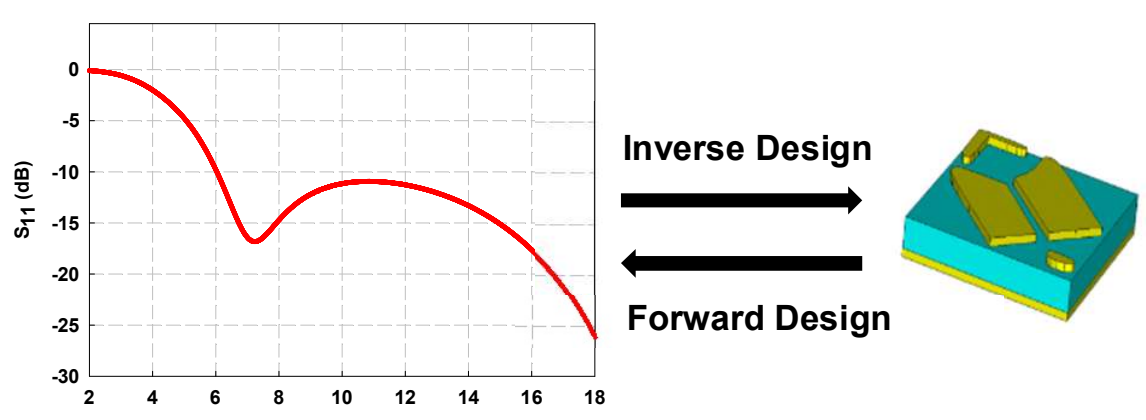

Fig. 1. Illustration of forward and inverse design of metasurface based absorbers

To overcome the limitations of traditional approaches, deep learning (DL) has emerged as a powerful alternative for inverse design of metasurfaces [32], [33]. DL techniques can efficiently explore high-dimensional design spaces and rapidly generate meta-atom configurations that satisfy specific EM performance criteria. Once trained, DL models offer near-instantaneous predictions, eliminating the need for iterative full-wave simulations for every new design target. Initial approaches focused on predicting the structural parameters of predefined geometries by training on datasets composed of explicit input–output pairs, where each design parameter set is linked to its corresponding EM response. Through iterative training, the network learns to infer suitable parameter combinations that achieve specific EM responses [34]-[36]. Several models adopted a tandem architecture, coupling an inverse neural network with a forward network to address the intrinsic non-uniqueness of the inverse problem and ensure physical consistency between predicted geometries and EM responses [37]-[39]. However, these early methods have serious limitations, especially for the design of free-form shapes, where high randomness and weak symmetry make it hard for standard networks to generalize well.

Extending beyond conventional DL approaches, generative models enable the generation of novel structures by learning and sampling from the underlying distributions of complex, high-dimensional datasets. Generative adversarial networks (GANs), particularly their conditional variants (cGANs), enable the generation of complex, free-form geometries conditioned on target EM responses [40]-[45]. To address typical GAN training problems such as mode collapse and unstable convergence, researchers have also adopted the Wasserstein GAN with gradient penalty (WGAN-GP), which provides more stable training and helps maintain diversity in the generated metasurface designs [46]-[48]. Autoencoders offer an alternative generative framework for inverse design of metasurfaces by constructing latent spaces that enable design exploration [49]. Variational autoencoders (VAE) have been used to embed metasurface structures and their scattering responses into a low-dimensional latent space, where PSO has been applied to identify optimal latent variables and consequently the corresponding metasurface structure [50]-[52]. They have also been successfully applied to realize various metasurface functionalities, including absorber designs [53]-[55] and for generating user specific structural color [56]. Recent hybrid approaches even combined VAEs with quantum GANs for enhanced optimization [57].

However, it is to be noted that, GANs and VAEs face significant limitations in inverse design applications. Traditional VAEs suffer from characteristic limitations including blurred reconstructions due to the reconstruction loss balanced against KL divergence, limited diversity and the requirement for additional optimization algorithms in latent space adding further computational overhead to the design workflow. GANs, while capable of generating high-quality structures, suffer from fundamental instability caused by the adversarial training process, making consistent conditioning on performance requirements difficult. Although cGANs were developed to address conditioning challenges, they were restricted to categorical conditions. Hence, they do not perform well in engineering applications where conditions are performance metrics which are continuous variables. Although continuous conditional GANs were introduced to address continuous conditioning, their success has largely been confined to computer vision applications [58] and they have been designed to condition generation on scalar performance metrics [59]. Diffusion-based generative models have recently emerged as promising tools due to their ability to learn the progressive noise corruption process and recover high-quality samples through iterative denoising. These models have demonstrated superior generation quality and accuracy compared to GAN and VAE-based approaches across several application domains, including text-to-image synthesis, image super-resolution, and metasurfaces [60]-[61].

In this regard, the present work proposes a condition-quality enhanced diffusion framework for inverse design of metasurface-based radar absorbing structures (RAS) under continuous EM constraints. The proposed framework introduces a feature-wise linear modulation (FiLM) based conditioning mechanism that propagates spectral and material constraints throughout the denoising hierarchy, enabling stable generation with enhanced conditional alignment. To further ensure adherence to the desired EM response, a pre-trained surrogate EM simulator (SES) is integrated directly into the generative learning process, thereby embedding spectral consistency into training itself rather than performing post-generation physics validation. The framework takes frequency-dependent reflection characteristics and material properties as conditioning inputs, enabling fabrication-constrained generation using commercially available materials and feasible substrate thicknesses. Further, in contrast to most existing approaches that generate only the meta-atom geometry while keeping design factors such as pattern material fixed, the proposed model simultaneously predicts the complete metasurface configuration, including the pattern material and its corresponding resistivity. Furthermore, the framework addresses the inherently non-unique nature of inverse EM design by enabling generation of multiple geometrically distinct yet spectrally consistent meta-atoms for the same target response. The effectiveness of the proposed framework has been demonstrated for diverse reflection characteristics over the broad frequency range of 2–18 GHz. To validate practical realizability, a planar RAS based on the generated meta-atom has been fabricated and experimentally characterized, and the measured response has been compared with the simulated and target spectra.

This paper is organized as follows. Section II presents the theoretical background of conditional diffusion models. The dataset and encoding scheme are described in Section III. The overall architecture of the proposed framework is presented in Section IV. Section V demonstrates the capabilities of the proposed approach through representative design examples. Section VI presents the results of ablation analysis followed by Section VII presenting details of experimental validation. Finally, conclusions are drawn in Section VIII.

## II. Conditional Diffusion Models

Diffusion models are generative frameworks based on two complementary stages (Fig. 2): a forward diffusion process ($q$) and a reverse denoising process ($p$). The forward process ($q$) gradually adds noise to the original image $x_0$ in $T$ steps following a predefined variance schedule $\beta_1, \beta_2, \ldots, \beta_t$ as given below [62]:

$$q(x_t \mid x_{t-1}) = N(x_t; \sqrt{(1-\beta_t)} \cdot x_{t-1}, \beta_t \cdot I) \tag{1}$$

$$q(x_{1:T} \mid x_0) = \prod_{t=1}^{T} q(x_t \mid x_{t-1}) \tag{2}$$

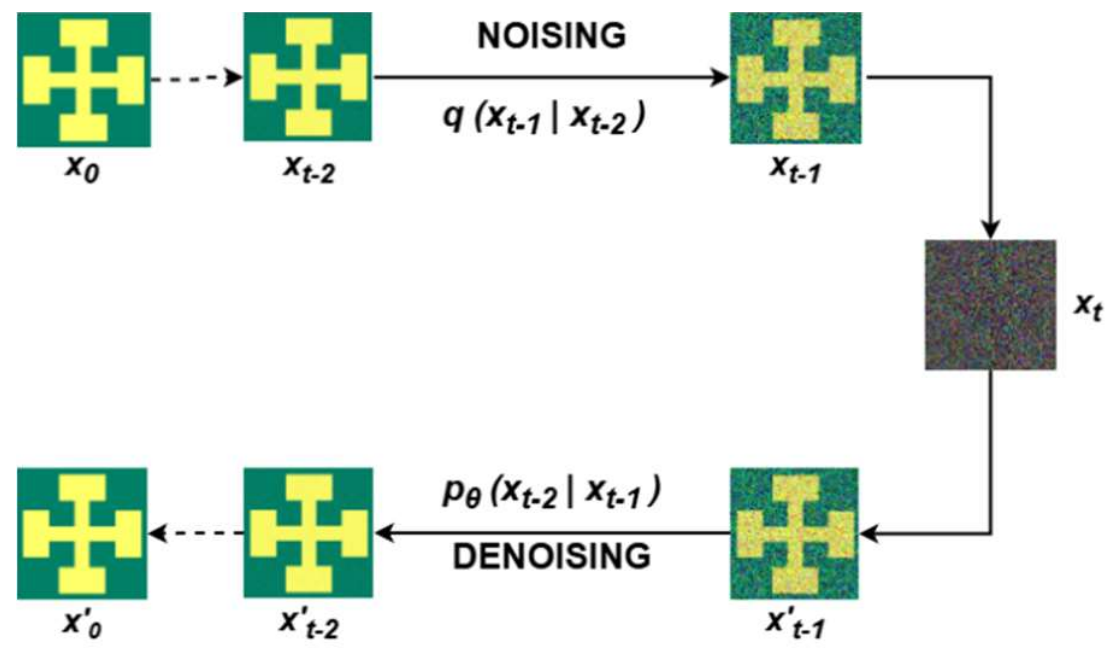

Fig. 2. Forward and backward processes in a diffusion model.

In the present context, $x_0$ denotes the image representing the meta-atom. The variance parameter $\beta_t$ increases with each timestep ($t$), resulting in progressively higher levels of noise being added to the image. Through this iterative noise injection process, the image gradually converge towards a Gaussian distribution.

By applying a re-parameterization technique, the corrupted image ($x_t$) at any arbitrary time step $t$ can be expressed directly in terms of the original image $x_0$ through a closed-form relationship as given below:

$$x_t = \sqrt{\bar{\alpha}_t}\, x_0 + \varepsilon\sqrt{1-\bar{\alpha}_t}\,, \text{ where } \varepsilon \sim N(0,1) \tag{3}$$

Where, $\alpha_t = 1 - \beta_t$ and $\bar{\alpha}_t = \prod_{s=1}^{t} \alpha_s$

The reverse denoising process ($p$) constitutes the core generative mechanism wherein the model learns to systematically remove noise and reconstruct the original data distribution. It can be represented as [62],

$$p_\theta(x_{t-1} \mid x_t) = N\left(x_{t-1};\, \mu_\theta(x_t, t),\, \textstyle\sum_\theta (x_t, t)\right) \tag{4}$$

During training, $p_\theta(x_{t-1}|x_t)$ should become as close as possible to $q(x_{t-1}|x_t)$. This problem is equivalent to training a neural network $\varepsilon_\theta(x_t, t)$ to predict the noise $\varepsilon_t$ that was added at each time step during the forward process, i.e.

$$\min_\theta L(\theta) = \mathrm{E}_{\{x_0,\varepsilon,t\}} \left[ \| \varepsilon_t - \varepsilon_\theta(x_t, t) \|^2 \right] \tag{5}$$

where $\theta$ represents the model parameters, and $x_t$ can be obtained from (3).

Further, in order to steer the generation of images towards a specific condition, a guidance mechanism is required. In the present work, classifier-free guidance (CFG) is employed to achieve conditional generation by jointly training the model on both conditional and unconditional objectives [63]. The neural network $\varepsilon_\theta$ is now conditioned on $c$, denoted as $\varepsilon_\theta(x_t, t, c)$. During training, the model is presented with the conditioning vector $c$ for most of the training steps. However, for a fixed percentage of steps, $c$ is replaced with a null token $\phi$, forcing the model to learn the unconditional generation path as well. During inference, the model makes two predictions at each time step $t$: conditional noise prediction $\varepsilon_\theta(x_t, t, c)$and unconditional noise prediction $\varepsilon_\theta(x_t, t, \phi)$. The final predicted noise $\hat{\varepsilon}_\theta$ is then computed as:

$$\hat{\varepsilon}_\theta(x_t,t,c) = \varepsilon_\theta(x_t,t,c) + w\left(\varepsilon_\theta(x_t,t,c) - \varepsilon_\theta(x_t,t,\phi)\right) \tag{6}$$

Where $w$ is the guidance scale that controls the extent of conditioning. The training objective remains the same as in equation (5), but is now applied to the conditional model. The guided noise prediction $\hat{\varepsilon}_\theta$ calculated in equation (6) is used in place of $\varepsilon_\theta$ in equation (5) to generate the sample for the previous timestep, $x_{t-1}$ .

## III. Description of Dataset & Encoding Scheme

To facilitate effective training of the diffusion-based generative model, a dataset comprising 39,664 samples across 30 geometrically distinct classes of meta-atoms has been developed. The parameter combinations for each class have been systematically selected, accounting for fabrication tolerances, to generate a wide range of reflection characteristics in the frequency range of 2–18 GHz. To capture both material and configuration dependent EM variations, each meta-atom has been simulated using various commercially available substrate materials with feasible thicknesses, different pattern (metallic/resistive) materials, and in both single-layer and dual-layer configurations. Dual-layer configurations are incorporated to represent practical scenarios where the patterned layer requires protection under harsh environmental conditions. The configurations of RAS geometries included in the dataset are illustrated in Fig. 3.

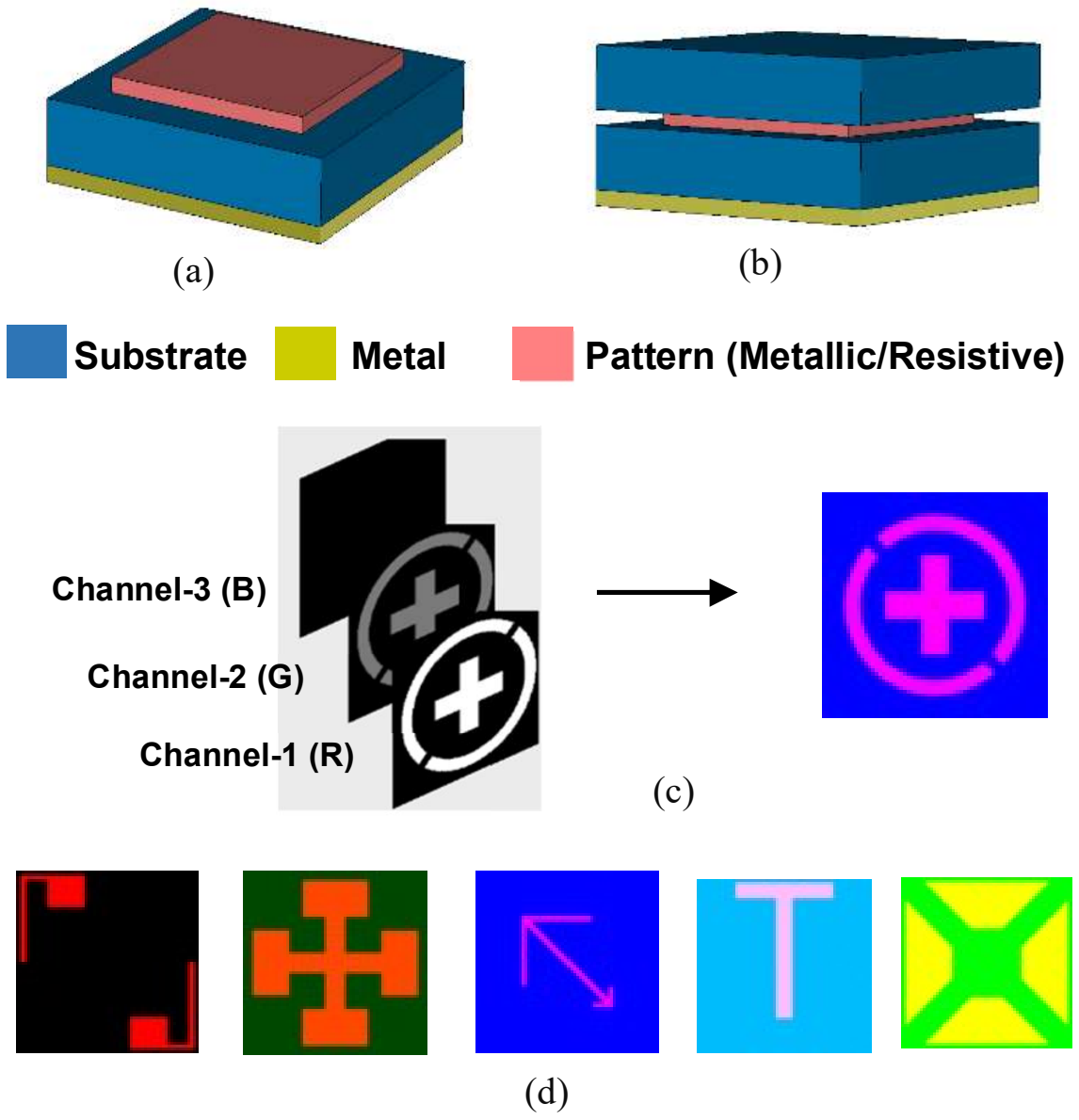

Fig. 3. Configuration of metasurface based RAS included in dataset (a) Single-layered RAS (b) Dual-layered RAS (c) Encoding scheme (d) Samples of encoded meta-atoms.

The dataset has been generated using full-wave EM simulations in CST Studio Suite [65]. The meta-atoms have been analyzed using frequency-domain solver with periodic boundary conditions along the lateral directions, open (add space) boundaries normal to the surface, and plane-wave excitation to model the incident EM field.

On generation of data, the configuration of meta atom has been encoded as a 3-channel colored image. The first channel (R) represents the geometric pattern of the meta-atom. Here, a pixel value of 255 indicates the presence of metallic or resistive material and 0 denotes its absence. The second channel (G) contains the same geometric pattern weighted by the normalized resistivity of the pattern material. The third channel (B) encodes the layer configuration by weighting the pattern with a binary value of 0 or 1 to represent single-layered and dual-layered meta-atoms, respectively. The resulting multi-channel representation of the meta-atom is illustrated in Fig. 3c. The proposed RGB encoding scheme uniquely represents the design space, effectively reducing the extent of one-to-many mapping between EM responses and metasurface geometries. The associated conditioning input consists of the reflection characteristics represented by $|S_{11}|$ sampled at 201 frequency points in the frequency range of 2–18 GHz and the substrate material parameters including relative permittivity ($\varepsilon_r$), dielectric loss tangent ($\tan \delta_e$), and substrate thickness. The material properties are incorporated as conditioning inputs to constrain the generative process to manufacturable metasurface configurations using commercially available materials and feasible thickness ranges.

## IV. Architecture of the Network

The architecture of the proposed framework, illustrated in Fig. 4, is designed to synthesize the complete configuration of meta-atom capable of achieving specified reflection characteristics for a given substrate material. The main components of the proposed framework are described below:

(i) *Condition embedding network*: Since the conditioning input consists of different physical attributes i.e., reflection characteristics and material parameters, they are processed through separate embedding networks to extract meaningful feature representations. The resulting embeddings are then concatenated to form a unified condition vector ($c$) as given below,

$$c=[e_{S_{11}}, e_m]$$

where $e_{S_{11}}$ and $e_m$ denote the spectral and material embeddings, respectively. This combined embedding serves as the conditioning input to the diffusion model.

(ii) *Conditional diffusion-based generative network*: The proposed framework employs a conditional diffusion model as the generative backbone, following the standard diffusion paradigm described in Section II. During training, the forward diffusion process progressively corrupts the images representing meta-atoms with Gaussian noise over multiple timesteps. The reverse process is modeled using a U-Net based denoising network (Fig. 4d) comprising an encoder–decoder architecture with residual convolutional blocks and skip connections for preserving fine geometric details. The network takes the noisy image, diffusion timestep, and conditioning information as inputs, and learns to predict the injected noise, thereby reconstructing the underlying image of meta-atom. Classifier-free guidance, as discussed in Section II, is incorporated to enhance conditional controllability. During inference, the model iteratively denoises an initial Gaussian noise sample through the reverse diffusion process to synthesize meta-atoms satisfying the specified conditioning constraints.

(iii) *Conditioning via FiLM Modulation*: To enable stable and accurate generation of meta-atoms under continuous EM constraints, we introduce a FiLM-based conditioning mechanism within the residual blocks of the denoising U-Net, as illustrated in Fig. 4(c). Unlike categorical conditioning, inverse design under continuous spectral constraints requires effective propagation of condition information across the denoising hierarchy to maintain spectral fidelity throughout the generation process. To achieve this, the embedded condition vector ($c$) is processed using a multilayer perceptron (MLP) to generate modulation parameters ($\gamma$, $\beta$). These parameters perform feature-wise affine modulation of intermediate activations through scaling and shifting operations, enabling the network to adapt its internal representations according to the desired EM response. The FiLM operation as shown in Fig. 4b can be expressed as:

$$FiLM\ (F) = \gamma(c)F + \beta(c) \tag{7}$$

Where $\gamma(c)$ and $\beta(c)$ are learned functions of $c$ and $F$ denote the feature map of a layer.

(iv) *Surrogate-based physics integration*: To embed EM consistency directly into the generative learning process, a pre-trained surrogate EM simulator is integrated within the diffusion training pipeline as shown in Fig. 4a. Unlike conventional approaches, the proposed framework enforces spectral consistency during training itself, thereby constraining the generative process using EM principles. The SES maps the generated meta-atom $x_0$ to its corresponding reflection spectrum ($S_{11\,(gen)}$), which is then compared with the target spectrum ($S_{11\,(target)}$) to compute a physics-guided spectral consistency loss ($L_{spec}$) as given below:

$$L_{spec} = \sum_{i=1}^{N_f} \left( \left| S_{11(gen)}\left(f_i\right) \right| - \left| S_{11(target)}\left(f_i\right) \right| \right)^2 \tag{8}$$

Where $N_f$ refer to the number of frequency ($f$) points. This loss enables the diffusion model to generate metasurface geometries that satisfy the desired EM performance constraints.

The overall loss function of the proposed inverse design framework has been then formulated as given below:

$$L = w_R L_R + w_G L_G + w_B L_B + \lambda L_{spec} \tag{9}$$

Where $L_R$, $L_G$ and $L_B$ denote the diffusion loss as given in equation (5) computed across the three channels of the generated image of meta-atom. $w_R$, $w_G$, $w_B$ and $\lambda$ are hyper parameters controlling the influence of different loss components during training.

## V. Results and Discussion

This section evaluates the performance of the proposed inverse design framework in terms of conditional alignment of the generated meta-atoms. Prior to model training, a comprehensive analysis of the spectral data distribution was performed to characterize the diversity of EM responses present in the dataset. The analysis revealed a significant imbalance across spectral categories, with certain response characteristics dominating the dataset. This can introduce bias into the learning process. To address this, the dataset was categorized into different spectral response classes, including single-resonance, multi-resonance, wideband, ultra-wideband characteristics, etc. Since the number of samples across different spectral categories is varied, the dataset has been partitioned into train, validation and test sets using a stratified category-wise splitting strategy with adaptive allocation. For categories with sufficient samples an 80:10:10 split was used, whereas categories with limited samples were split using a minimum-count strategy to ensure representation in validation and test sets without exhausting the training data. This approach guarantees representation of all spectral classes during evaluation while preserving sufficient training data for rare categories.

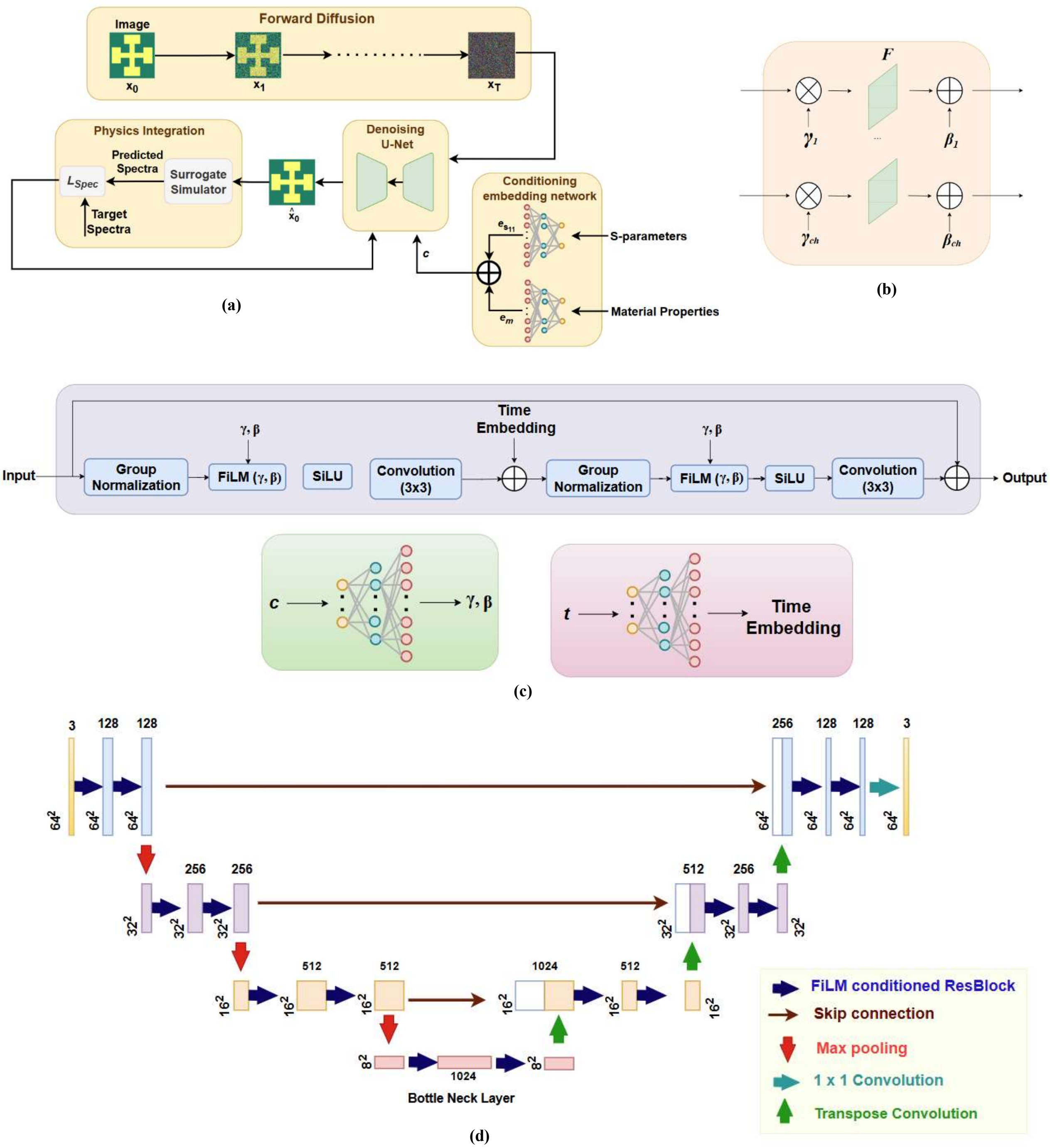

Fig. 4. Architecture of the proposed inverse design framework (a) Complete block diagram (b) A FiLM layer (*ch* denotes the number of channels in the feature map) (c) FiLM conditioned ResBlock (d) Denoising U-Net.

Furthermore, to improve learning of sparsely populated spectral categories, a weighted sampling strategy was incorporated during training. Here, each sample is assigned a weight inversely proportional to the number of samples in its corresponding spectral category. This enables balanced learning across diverse spectral responses and improved the robustness and generalization capability of the proposed framework. Subsequently, an extensive hyper parameter tuning has been performed and the following set of hyper parameters has been chosen: $w_R$ =2, $w_G$ = 1.5, $w_B$ = 1.5, $\lambda$ =5, $w$=5, initial learning rate= $10^{-4}$ and batch size =32. The model with the optimized set of hyper parameters has been trained for 400 epochs using Adam optimizer in conjunction with a cosine annealing learning rate schedule. Once trained, the model has been used to generate meta-atoms catering to specific conditional constraints. The data points in the test dataset have been used for this purpose. Fig. 5 illustrates the progressive evolution of the generated metasurface design from random noise to the final geometry during the iterative denoising process.

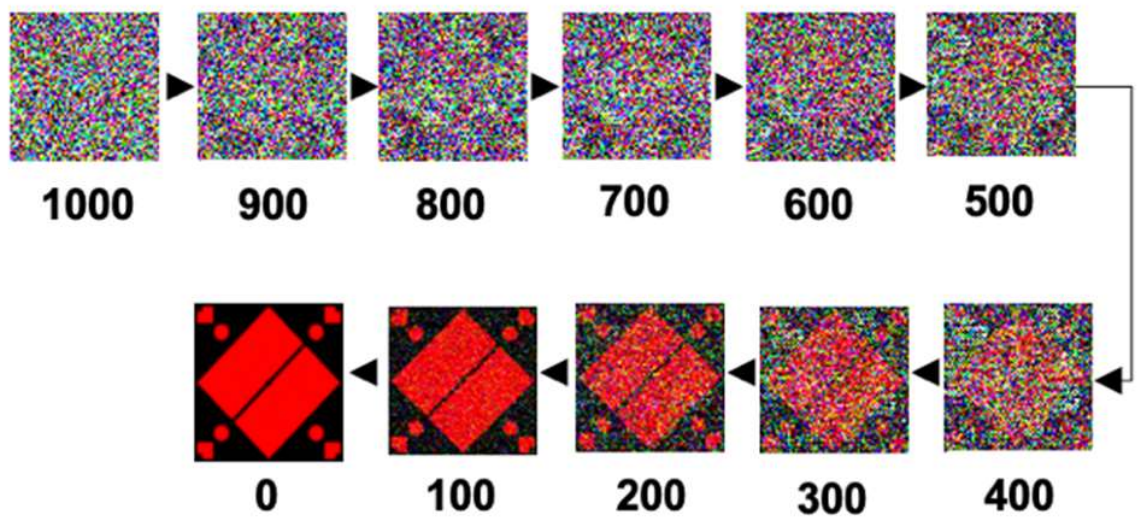

Fig. 5. Evolution of the meta-atom geometry from random noise during the iterative de-noising process at different time steps ($t$).

Fig. 6 shows the comparison between the target spectra and the spectra produced on simulation of the generated meta-atoms in CST. The images of the generated meta-atom patterns are shown as inset. The comparative plots indicate that the model is capable of generating meta-atoms that can produce a variety of spectral behaviors, including narrowband, multi-resonance, and broadband reflection characteristics. From Fig. 6c, it is evident that the proposed model can even generate meta-atoms satisfying challenging constraints such as ultra-wideband characteristics (bandwidth>10 GHz), despite the limited availability of such samples in the dataset. This capability can be attributed to the proposed weighted sampling strategy, which improves the learning of sparsely populated spectral regions during training. Further, the proposed model also demonstrated the capability to generate hybrid meta-atom geometries (Fig. 6a, Fig. 6b and Fig. 6e) by effectively combining structural features from multiple unit-cell patterns present in the training dataset, indicating strong feature-level generalization and interpolation capability. Further, the close agreement between the generated and target spectra demonstrates the effectiveness of the FiLM-based conditioning mechanism and physics guidance via SES. The generated designs also exhibit well-defined structural features that closely resemble realistic meta-atoms, with feature dimensions remaining within practical fabrication limits. This demonstrates the effectiveness of the dataset, which was constructed while incorporating fabrication constraints to ensure the generation of physically realizable metasurface geometries.

Furthermore, the ability of the model to produce multiple geometrically distinct solutions corresponding to the same spectral requirement is shown in Fig. 7.

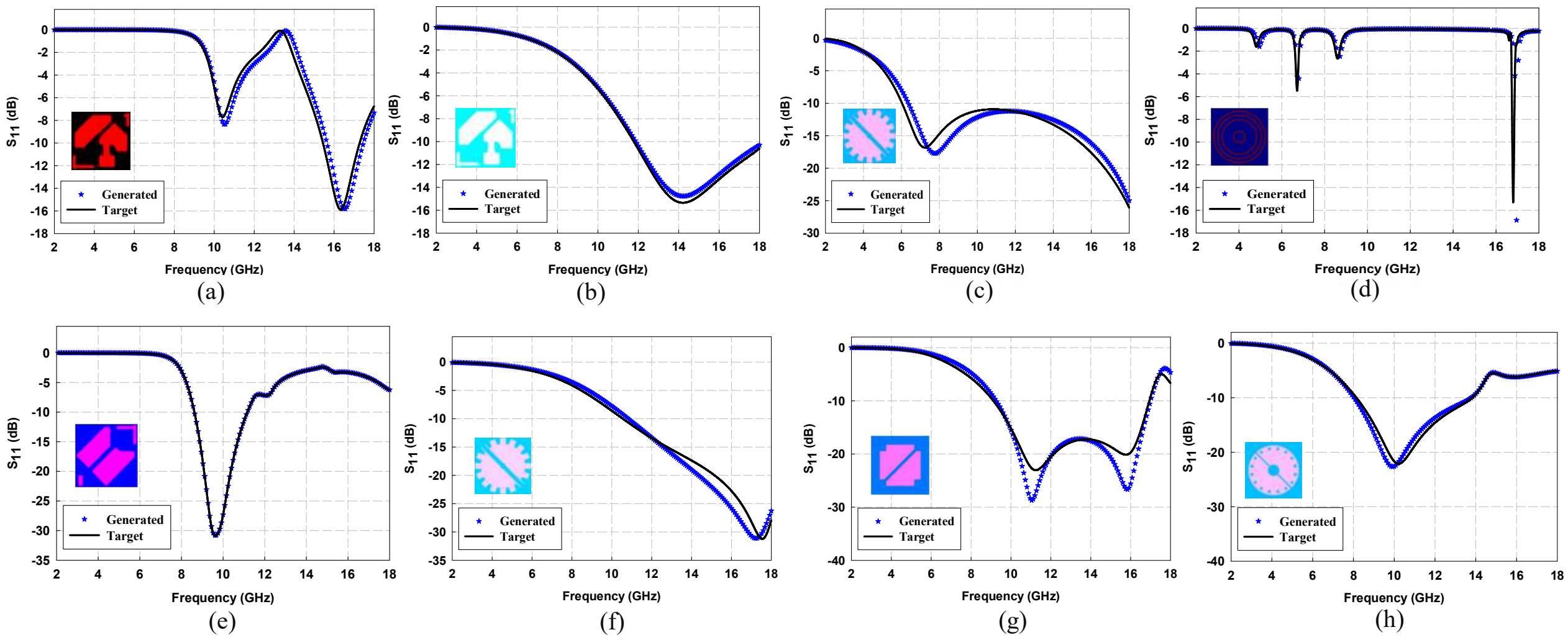

Fig. 6. Comparison between target spectra and the spectra obtained on simulation of generated meta-atoms for different samples in the test set. (a) Sample-1 (Material: RT/Duroid 5880 ($\varepsilon_r$=2.2, tan$\delta$=0.0009, $t$=1.575mm); Single-layered; Metallic pattern) (b) Sample-2 (Material: RO4835 ($\varepsilon_r$=3.48, tan$\delta$=0.0037, $t$=1.524mm); Double-layered; Resistive pattern (100 ohm/sq.)) (c) Sample-3 (Material: AD255C ($\varepsilon_r$=2.6, tan$\delta$=0.0013, $t$=3.175mm); Double-layered; Resistive pattern (70 ohm/sq.)) (d) Sample-4 (Material: RO4835 ($\varepsilon_r$=3.48, tan$\delta$=0.0037, $t$=1.524mm); Double-layered; Metallic pattern) (e) Sample-5 (Material: RO4835 ($\varepsilon_r$=3.48, tan$\delta$=0.0037, $t$=1.524mm); Double-layered; Metallic pattern) (f) Sample-6 (Material: RO4533 ($\varepsilon_r$=3.45, tan$\delta$=0.0025, $t$=1.524mm); Double-layered; Resistive pattern (75 ohm/sq.)) (g) Sample-7 (Material: Kappa 438 ($\varepsilon_r$=4.38, tan$\delta$=0.005, $t$=1.524mm); Double-layered; Resistive pattern (50 ohm/sq.)) (h) Sample-8 (Material: RO4360G2 ($\varepsilon_r$ =6.4, tan$\delta$=0.0038, $t$=1.524mm); Double-layered; Resistive pattern (75ohm/sq)).

The diversity of the generated meta-atom geometries has been quantified using the following average pairwise structural dissimilarity metric ($D$):

$$D = \frac{2}{N(N-1)} \sum_{i<j} \left[\lambda d_1 + (1-\lambda) d_2 \right] \quad (10)$$

Where, $d_1 = 1 - \frac{|M_i \cap M_j|}{|M_i \cup M_j| + \epsilon}$; $d_2 = 1 - \frac{\left|R_i^{(\delta)} \cap R_j^{(\delta)}\right|}{\left|R_i^{(\delta)} \cap R_j^{(\delta)}\right| + \epsilon}$

$M_i$ and $M_j$ denote the binary masks corresponding to the $i^{th}$ and $j^{th}$ generated meta-atom geometries, respectively, where 1 represents the presence of pattern material and 0 represents the absence of it. $d_1$ measures the variations in pattern material distribution between a pair of images using the intersection over union (IoU) distance between binary masks. $R_i^{(\delta)}$ and $R_j^{(\delta)}$ denote the boundary regions of the corresponding masks, obtained by selecting pixels lying within a distance $\delta$ from the metal edges. $d_2$ quantifies fine-scale geometric differences by comparing the contour regions. The weighting factor λ controls the relative contribution of global structural variation and local boundary variation. A small constant ∈ is introduced in the denominator to ensure numerical stability. $N$ denotes the number of generated images. The diversity metric yielded a high score of 0.803 for the proposed model. Further, the generated samples exhibited high spectral consistency with a very low average mean squared error of 0.001 under identical conditioning. Such diversity is particularly beneficial as it provides designers with multiple candidate geometries allowing flexibility in fabrication and implementation.

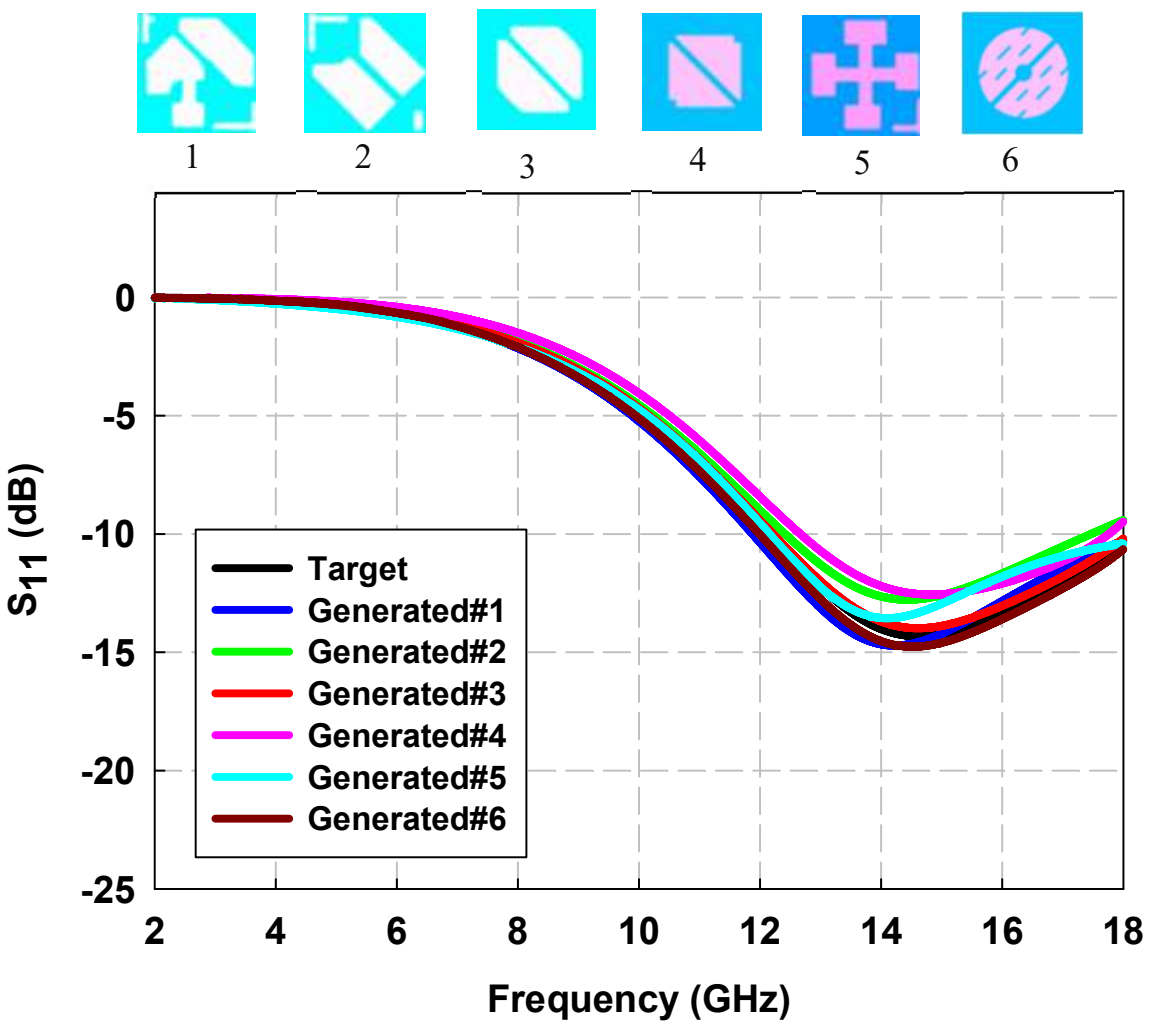

Fig. 7. Diverse metasurface geometries generated by the proposed model for the same target reflection characteristics.

## VI. Ablation Analysis

This section presents an ablation analysis to assess the effectiveness of diffusion-based models for inverse design of metasurface-based RAS relative to other generative frameworks such as conditional WGAN-GP and VAE, along with a systematic assessment of different conditioning mechanisms. The performance of each variant is evaluated using multiple metrics including average spectral mean squared error (MSE), accumulated average error (AAE), average band alignment accuracy (BAA), and the percentage of valid designs. All evaluation metrics are computed over the test dataset. The average spectral MSE and AAE quantify pointwise agreement between the target and generated spectra. They are defined as:

$$MSE = \frac{1}{N_{test}} \sum_{j=1}^{N_{test}} \frac{1}{N_f} \sum_{i=1}^{N_f} \left( \left|s_{11(gen)}(f_i)\right| - \left|s_{11(target)}(f_i)\right| \right)^2 \quad (11)$$

$$AAE = \frac{1}{N_{test}} \sum_{j=1}^{N_{test}} \frac{1}{N_f} \sum_{i=1}^{N_f} \left\| s_{11(gen)}(f_i) \right| - \left| s_{11(target)}(f_i) \right\| \quad (12)$$

Where $|S_{11(gen)}(f_i)|$ and $|S_{11(target)}(f_i)|$ denote the generated and target spectral response at the $i^{th}$ frequency point ($f_i$). $N_f$ denotes total number of frequency points and $N_{test}$ denote number of data points in the test data set.

BAA measures the overlap between the desired and generated absorption regions and is defined as,

$$BAA = \frac{1}{N_{test}} \sum_{j=1}^{N_{test}} \frac{\left|B_{target}(j) \cap B_{gen}(j)\right|}{\left|B_{target}(j)\right|} \quad (13)$$

where $B_{target}$ and $B_{gen}$ refer to the set of frequency points in the target spectrum and generated spectrum respectively where $|S_{11}| \leq$ -10dB. The percentage of valid designs denote the fraction of generated designs that has a band alignment greater than 0.8. Table I presents the summary of the ablation analysis *w.r.t.* the above-mentioned evaluation metrics.

**Table I** Summary of ablation analysis

| Model | Avg. MSE | AAE | BAA | Valid designs |
|---|---|---|---|---|
| WGAN-GP | 0.049 | 0.168 | 0.302 | 30% |
| Baseline diffusion | 0.045 | 0.140 | 0.483 | 40% |
| Baseline diffusion with SES and condition concatenated with image at input | 0.042 | 0.132 | 0.578 | 60% |
| Baseline diffusion with SES and condition injection via FiLM in U-Net (Proposed) | 0.0006 | 0.016 | 0.958 | 100% |

It is clear from Table I that the baseline diffusion model is better compared to conditional WGAN-GP *w.r.t* different metrics. The geometries generated by VAE exhibited poor structural quality, characterized by numerous irregular fine-scale features that were difficult to simulate in CST and impractical for fabrication. Hence, the corresponding metrics are not included. Further, it is evident that the progressive inclusion of conditioning mechanisms leads to consistent improvement across all performance metrics. In particular, the integration of SES and FiLM-based conditioning significantly enhances spectral accuracy and band alignment. Overall, these results underscore the importance of combining physics-guided learning with feature-wise conditioning for robust inverse metasurface design. In addition, it is to be noted that the proposed model generates the suitable design for meta-atom in approximately 30 seconds for a given target response, whereas conventional design approaches based on iterative full-wave simulations may require several months under comparable computational resources.

## VII. Experimental Validation

This section presents the practical feasibility of the proposed inverse design framework via experimental validation. The proposed model has been used to predict the meta-atom geometry for a specific reflection spectrum as shown in Fig. 8a to be realized using RO4835 substrate material ($\varepsilon_r$=3.48, tan$\delta$=0.0037, $t$=1.524mm). The generated meta-atom (Fig. 8b) has been used to configure a 15cm×15cm planar RAS in CST Studio suite as shown in Fig. 8c and Fig. 8d.

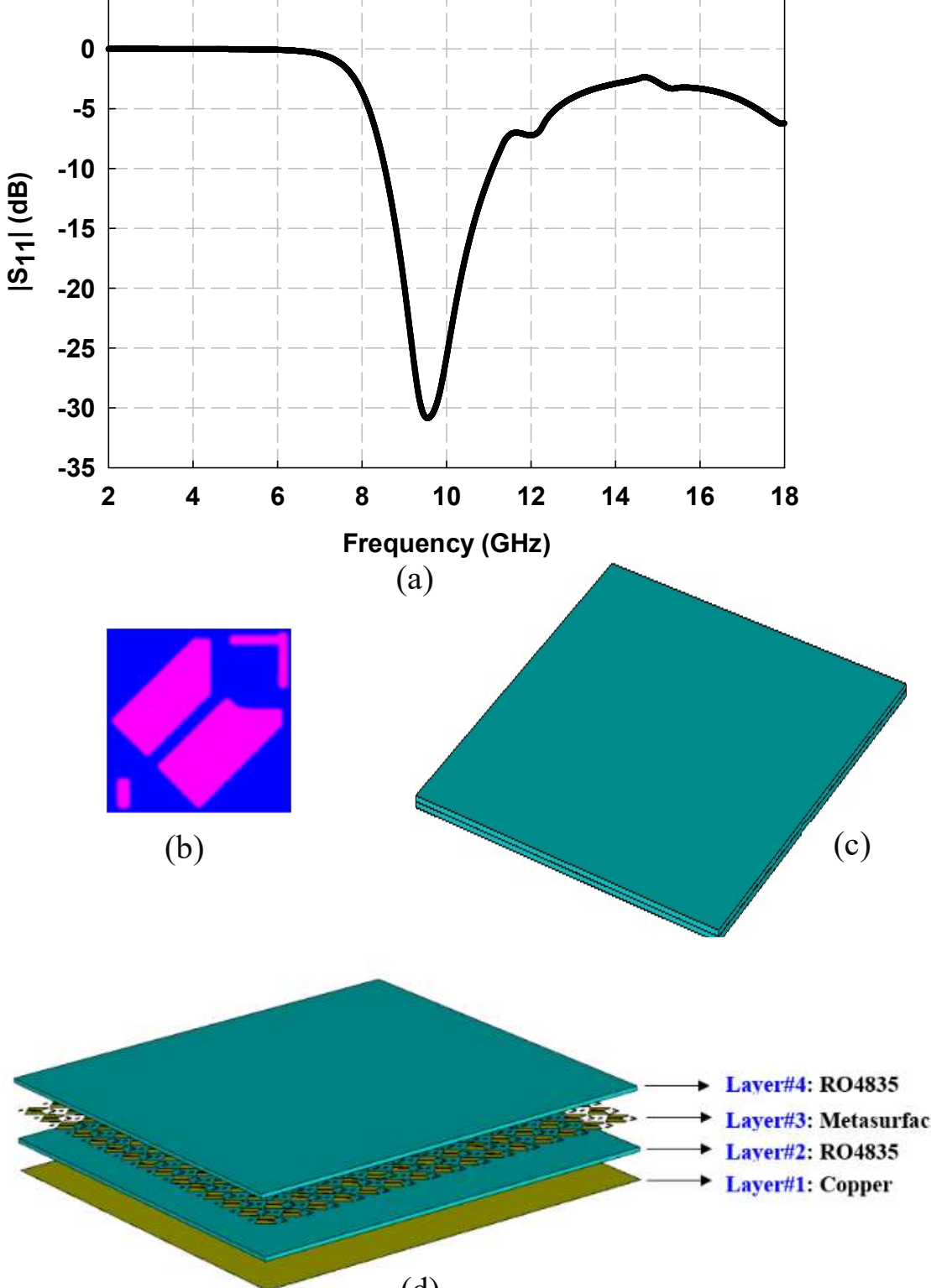

Fig. 8. (a) Target reflection spectra (b) Meta atom geometry generated using the proposed inverse design framework (c) Simulated model of planar RAS (15cm x 15cm) (d) Exploded side view of simulated planar RAS.

The planar RAS has been checked for fabrication feasibility by confirming that the minimum feature size is greater than 0.1mm. Once confirmed via simulation, the same planar RAS has been fabricated via laser etching procedure. The fabricated prototype is shown in Fig. 9.

Since the target spectra has $S_{11}$ ≤ -10dB within the frequency range of 8.5 GHz to 11 GHz, the reflection characteristics ($S_{11}$) of the fabricated planar RAS under normal incidence in this frequency range has been measured using a free-space measurement setup consisting of an Agilent Technologies PNA Network Analyzer (N5227A) and a pair of X-band horn antennas. The measurement set-up is shown in Fig. 10a. The absorption performance of the structure has been then evaluated based on the following relation:

$$|A|^2 = 1 - |S_{11}|^2 - |S_{21}|^2 \quad (14)$$

where $|A|^2$ , $|S_{11}|^2$ and $|S_{21}|^2$ denote the absorbed, reflected, and transmitted power, respectively. In the present configuration, the transmission term $|S_{21}|^2$ is negligible due to the presence of the continuous metallic backing, resulting in $|A|^2 \approx 1 - |S_{11}|^2$.

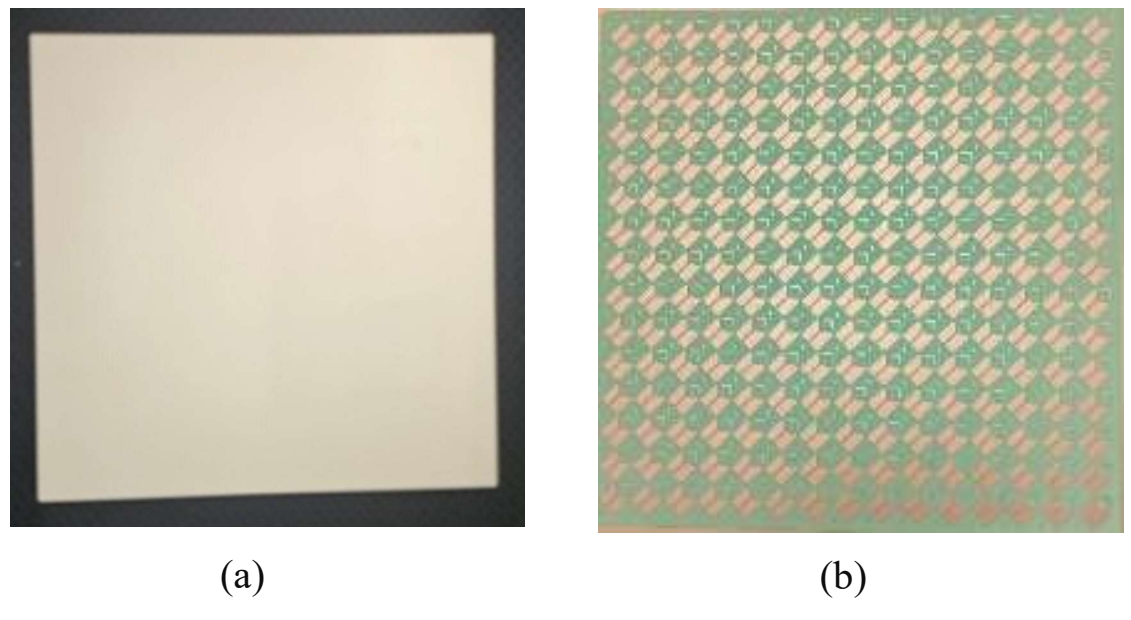

(a) (b)

Fig. 9. Fabricated prototype of planar RAS (15cm x 15cm) (a) Top view (b) Etched metasurface pattern (green color denote the photoresist layer).

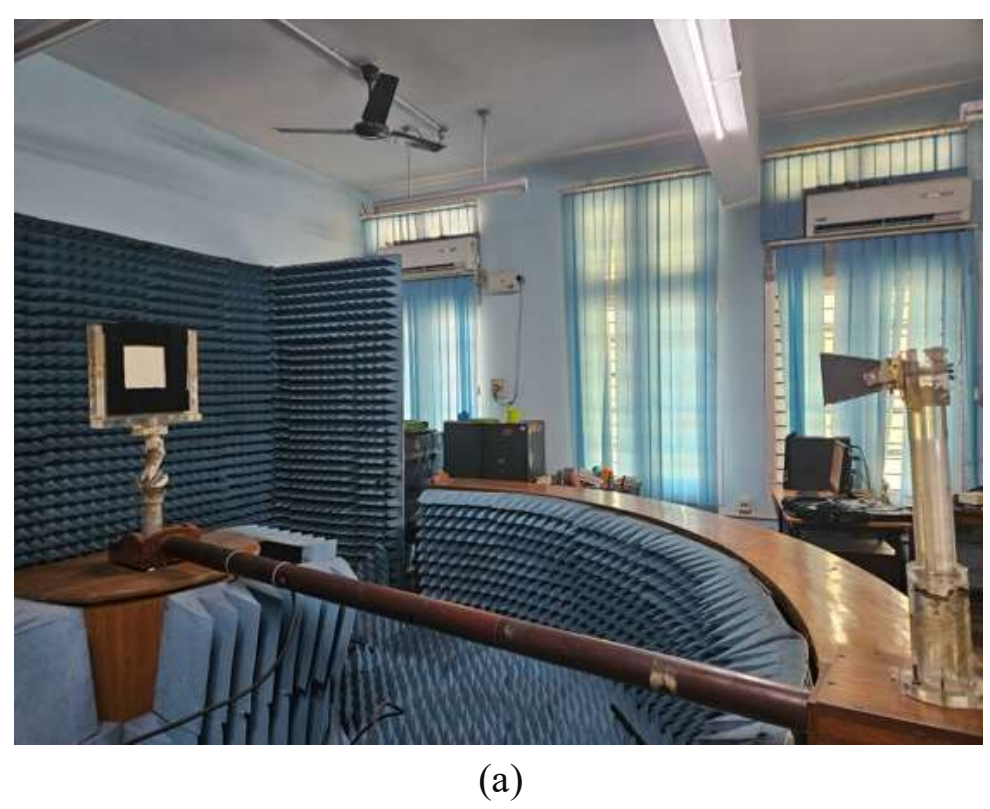

(a)

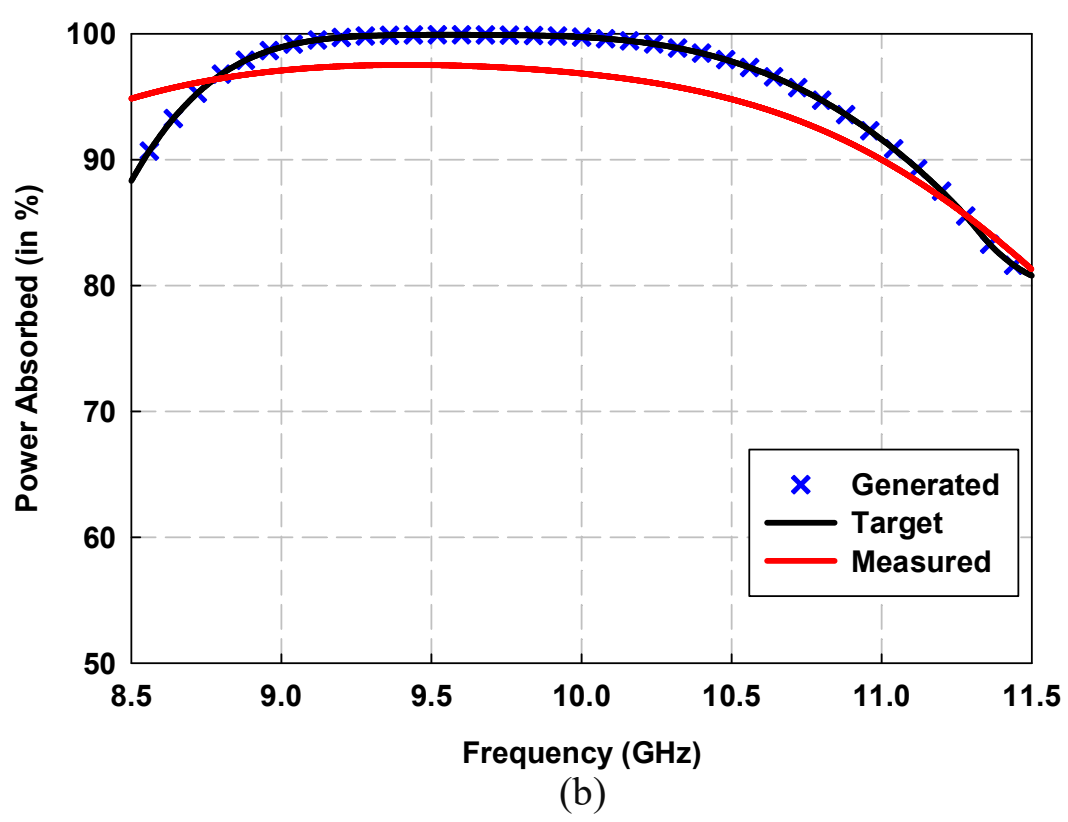

(b)

Fig. 10. (a) Experimental set up for measurement of $S_{11}$ (b) Target, generated and measured absorption characteristics of RAS.

The frequency-dependent absorption characteristics obtained from measurements have been compared with the generated results and the target spectrum in Fig. 10b. It can be observed that the fabricated RAS achieves greater than 90% power absorption within the targeted frequency range. Minor deviations can be attributed to fabrication tolerances and measurement uncertainties. This agreement confirms the capability of the proposed framework to accurately translate the desired EM response into a realizable metasurface design. Table II presents the comparison of existing generative techniques and the proposed approach *w.r.t.* inverse design of metasurfaces.

**Table II** Comparison of proposed approach with prior art

| Ref. | Freq. range | Model | Size of dataset | Condition | Physics Integration | Conditioning Quality Assessment | | | Diversity | Exp. validation |
|---|---|---|---|---|---|---|---|---|---|---|
| | | | | | | Qualitative | Quantitative | Scenarios | | |
| [40] | 170 THz-600 THz | GAN | 8000 | Transmittance spectra | Yes | Yes | No | Limited | No | No |
| [41] | 8 GHz - 12 GHz | GAN | 54000 | Scattering parameters | No | Yes | No | Limited | No | No |
| [42] | 7 GHz - 20 GHz | GAN | 4700 | Axial ratio | No | Yes | No | Limited | No | Yes |
| [43] | 4 GHz - 20 GHz | GAN | 850 | Reflection spectra | No | Yes | No | One | No | No |
| [44] | 25 THz - 75 THz | GAN | 18768 | Absorption spectra | Yes | Yes | Yes | Limited | Yes | No |
| [47] | 20 GHz - 30 GHz | GAN | 12700 | Reflection spectra | Yes | Yes | Yes | Limited | No | No |
| [48] | 8 GHz - 12 GHz | GAN | 24000 | Transmission spectra | Yes | Yes | Yes | Limited | Yes | No |
| [50] | 15 GHz - 31 GHz | VAE | 19854 | Transmission spectra | Yes | Yes | No | One | No | No |
| [51] | 15 GHz - 31 GHz | VAE | 34000 | Transmission spectra | Yes | Yes | No | Limited | No | No |
| [52] | 5 GHz - 15 GHz | VAE | 3000 | Reflection spectra | No | Yes | Yes | Limited | No | No |
| [53] | 1 GHz - 18 GHz | VAE | 3116 | Scattering parameters | Yes | Yes | No | Limited | No | Yes |
| [54] | 2 GHz - 22 GHz | VAE | 13401 | Absorption spectra | No | Yes | No | Limited | No | Yes |
| [55] | 2 GHz - 40 GHz | VAE | 26000 | Scattering parameters | Yes | Yes | No | Limited | No | Yes |
| **Present work** | **2 GHz – 18GHz** | **Diffusion** | **39664** | **Reflection spectra** | **Yes** | **Yes** | **Yes** | **Wide** | **Yes** | **Yes** |

## VIII. CONCLUSION

This paper presented a physics-aware conditional diffusion framework for inverse design of metasurface-based RAS under continuous EM specifications and fabrication constraints. By integrating FiLM-based conditioning and surrogate-guided spectral supervision directly within the generative process, the proposed framework achieved accurate, controllable, and electromagnetically consistent synthesis of manufacturable metasurface configurations using commercially available materials and feasible thicknesses. The optimized framework achieved an average MSE of 0.0006, AAE of 0.016, diversity score of 0.803, band alignment accuracy of 0.958, and a valid EM design generation rate of 100% on the test dataset, demonstrating its capability for structured multimodal inverse design with high spectral fidelity. Experimental fabrication and measurements demonstrate good agreement between target, simulated, and measured responses, validating the practical realizability of the generated designs. In addition, the proposed approach reduces the metasurface design cycle from several months of iterative EM optimization to approximately 30s, significantly reducing reliance on computationally expensive full-wave simulations. Overall, the proposed framework bridges generative inverse design with realizable EM engineering constraints by enabling rapid synthesis of physically consistent, fabrication-aware metasurface configurations under continuous EM constraints. As future work, the approach can be extended to incorporate oblique angles of incidence and polarization dependent responses thereby further enhancing its applicability to real-world EM design problems.

## ACKNOWLEDGMENT

We express our gratitude to Council of Scientific and Industrial Research (CSIR), India for supporting this research activity. We also express our gratitude to CSIR-4PI for providing access to HPC facility, the Department of Electronics, Cochin University of Science and Technology (CUSAT) for supporting S-parameter measurements.

## DATA AVAILABILITY STATEMENT

The dataset supporting this study is not publicly available due to confidentiality restrictions associated with sponsored research projects involving proprietary metasurface designs.

## CONFLICT OF INTEREST

The authors declare that they have no conflict of interest.